\documentclass[letterpaper, 10 pt, journal, twoside]{IEEEtran}

\IEEEoverridecommandlockouts                              

\usepackage{times}
\usepackage{amsmath, amssymb, bm}
\usepackage{amsfonts}
\usepackage{graphicx}
\usepackage{subcaption}
\usepackage{comment}
\usepackage{algorithm}
\usepackage{algorithmic}
\usepackage{multicol}
\usepackage[bookmarks=true]{hyperref}
\usepackage{multirow}
\usepackage{xcolor}
\usepackage{acronym}
\usepackage[table]{xcolor}
\usepackage{bbm}
\usepackage{makecell}  
\usepackage{xcolor}

\title{\LARGE \bf
A Cross-Embodiment Gripper Benchmark for Rigid-Object Manipulation in Aerial and Industrial Robotics
}

\author{Marek Vagas$^{1}$, Martin Varga$^{1}$, Jaroslav Romancik$^{1}$, Ondrej Majercak$^{1}$, Alejandro Suarez$^{2}$, Anibal Ollero$^{2}$,Bram Vanderborght$^{3}$, and Ivan Virgala$^{1}$ 
\thanks{*This work is supported by the European ROBotics and AI Network (euROBIN, Grant agreement ID: 101070596) project funded by the European Commission}
\thanks{$^{1}$Marek Vagas, Martin Varga, Jaroslav Romancik, Ondrej Majercak and Ivan Virgala are with the Faculty of Mechanical Engineering, Technical University of Kosice, 04200 Kosice, Slovakia
        (emails:{marek.vagas}, martin.varga.2, jaroslav.romancik, ondrej.majercak, ivan.virgala, @tuke.sk)}%
\thanks{$^{2}$Alejandro Suarez and Anibal Ollero are with the GRVC Robotics Lab of Seville,
        41004 Sevilla, Spain
        (emails:{asuarezfm@us.es}, {aollero@us.es})}%
        \thanks{$^{3}$Bram Vanderborght is with with Brubotics,
Vrije Universiteit Brussel and IMEC,
        1050 Elsene, Belgium
        (email:{bram.vanderborhght@imec.be})}%
}

\begin{document}

\maketitle
\thispagestyle{empty}
\pagestyle{empty}


\begin{abstract}
Robotic grippers are increasingly deployed across industrial, collaborative, and aerial platforms, where each embodiment imposes distinct mechanical, energetic, and operational constraints. Established YCB and NIST benchmarks quantify grasp success, force, or timing on a single platform, but do not evaluate cross-embodiment transferability or energy-aware performance, capabilities essential for modern mobile and aerial manipulation.
This letter introduces the Cross-Embodiment Gripper Benchmark (CEGB), a compact and reproducible benchmarking suite extending YCB and selected NIST metrics with three additional components: a transfer-time benchmark measuring the practical effort required to exchange embodiments, an energy-consumption benchmark evaluating grasping and holding efficiency, and an intent-specific ideal payload assessment reflecting design-dependent operational capability. Together, these metrics characterize both grasp performance and the suitability of reusing a single gripper across heterogeneous robotic systems.
A lightweight self-locking gripper prototype is implemented as a reference case. Experiments demonstrate rapid embodiment transfer (median $\approx$ 17.6 s across user groups), low holding energy for gripper prototype ($\approx$ 1.5 J per 10 s), and consistent grasp performance with cycle times of 3.2–3.9 s and success rates exceeding 90\%. CEGB thus provides a reproducible foundation for cross-platform, energy-aware evaluation of grippers in aerial and manipulators domains.
\end{abstract}

\section{Introduction}

Robotic grasping has been extensively investigated across industrial, collaborative, and aerial domains. Over the past decades, numerous gripper designs have emerged. From simple parallel-jaw and underactuated mechanisms to modern soft, hybrid, and multifunctional systems ~\cite{citation_17}, ~\cite{citation_19}, ~\cite{citation_20}. Despite this progress, grippers deployed across heterogeneous platforms continue to face embodiment-dependent constraints such as mounting geometry, actuation compatibility, and power availability, which complicate direct comparison of their performance. Substantial technological progress, review papers consistently highlight the lack of a unified, quantitative 
\begin{figure}[!ht]
    \centerline{\includegraphics[width=7.5cm]{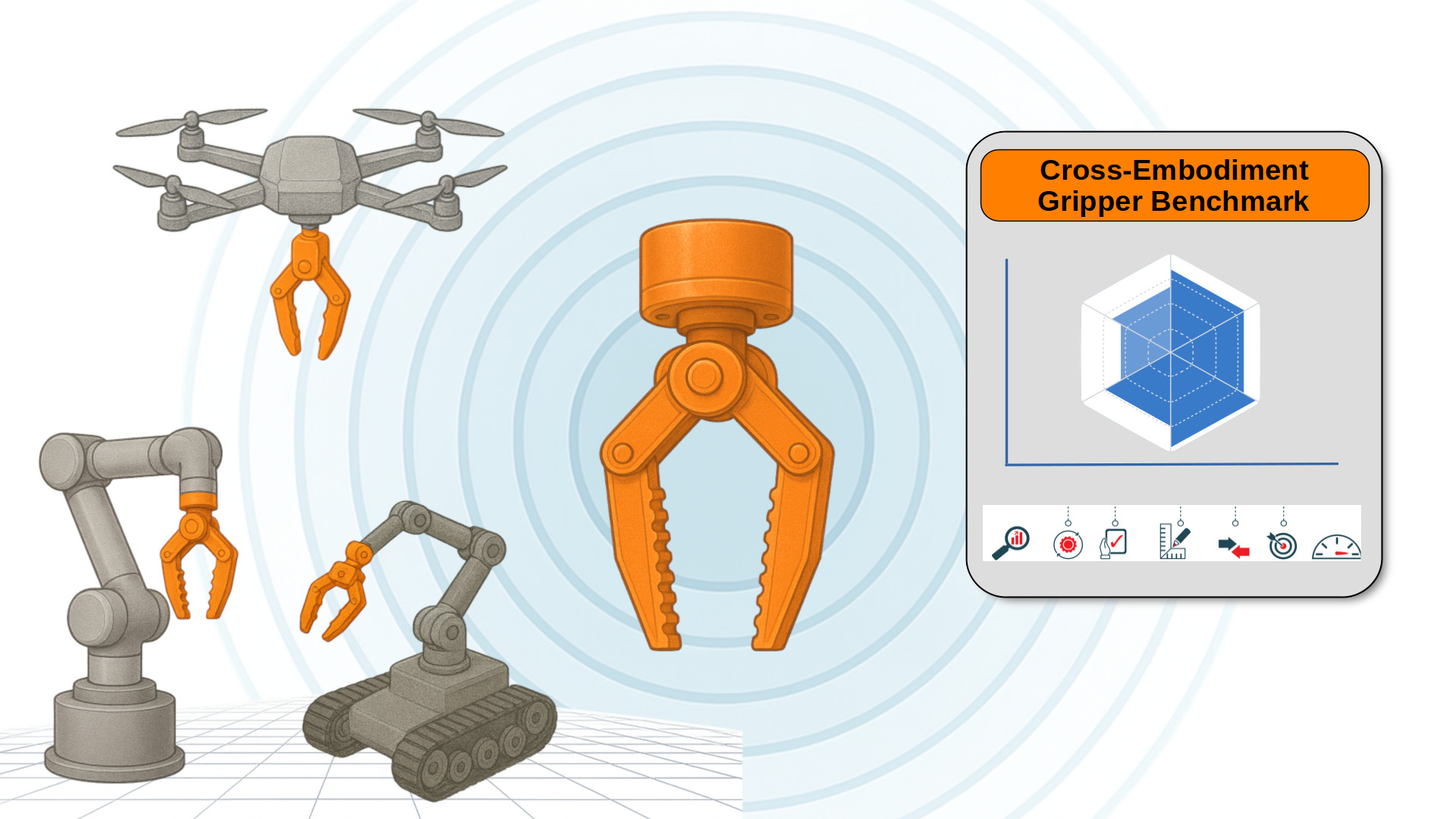}}
    \caption{Overview of the Cross-Embodiment Gripper Benchmark (CEGB). A common reference gripper is mounted on different robotic platforms—a collaborative manipulator, mobile robot with manipulator and UAV—to evaluate cross-embodiment transfer, mounting compatibility, and performance under different actuation and environmental conditions.}
    \label{figure 1}
\end{figure}
framework for cross-platform comparison ~\cite{citation_14}, ~\cite{citation_21}. Current benchmarks primarily quantify grasp success or adaptability but seldom address transferability, energy efficiency, or reproducibility across different robotic embodiments. To improve reproducibility, standardized object sets and test protocols have been introduced. The YCB Object and Model Set ~\cite{7251504} became a widely adopted reference for grasp success evaluation, while NIST protocols ~\cite{8950105}  
complemented it with timing, payload, and force-based metrics. Initiatives such as GRASPA ~\cite{citation_5}, OpenGRASP ~\cite{citation_2}, and the Robotic Grasping and Manipulation Competitions (RGMC) ~\cite{citation_10} have further advanced transparent benchmarking practices. However, these methods remain largely platform-specific and do not evaluate the ability of a single gripper to be reused across heterogeneous systems. 

The emergence of soft and hybrid grippers has emphasized the limitations of existing frameworks. Traditional benchmarks such as YCB and NIST were primarily designed for rigid or underactuated mechanisms, thus capturing only a subset of behaviors observed in compliant or multi-modal systems. Recent developments include multifunctional soft grippers with integrated sensing ~\cite{citation_27}, learning-based hybrid architectures ~\cite{citation_30}, and topology-optimized designs ~\cite{citation_26}. The SoGraB framework ~\cite{citation_11} introduced visual deformation analysis for soft grasping, representing a significant step forward. Nevertheless, these approaches focus mainly on shape adaptability and do not systematically consider energy efficiency or cross-platform transferability—factors critical for modern mobile and aerial robots. 

Aerial manipulation benchmarks ~\cite{citation_6} have addressed unique flight constraints such as payload, stability, and energy limitations, whereas industrial robotics including collaborative robots emphasize safety, modularity, and reconfigurability ~\cite{citation_16}, ~\cite{citation_15}. Modular and detachable grippers have therefore gained increasing attention, offering fast tool exchange and compatibility across multiple embodiments. Examples include the Co-Gripper ~\cite{Salvietti8593877}, a wireless modular end-effector capable of standalone operation, and detachable collaborative grippers using electromagnetic changers ~\cite{Iqbal} or modular fingertips ~\cite{Ringwald}. These concepts illustrate the emerging trend toward cross-platform reuse, yet their quantitative assessment remains largely unexplored. Although qualitative demonstrations repeatedly highlight the promise of such designs, the community lacks a standardized methodology for measuring how effectively a gripper can be transferred, redeployed, and operated across platforms with different kinematics, control strategies, and physical interfaces.

In the literature, transferability generally refers to the capability of a gripper to operate effectively across different embodiments, tasks, or environments ~\cite{citation_6}, ~\cite{Salvietti8593877}, ~\cite{Wang10582538}. Embodiment transfer describes mechanical or control adaptation between robotic platforms, task transfer concerns the reuse of grasp strategies across manipulation tasks, and environment transfer involves deploying the same gripper hardware across aerial and manipulators systems. While these notions are increasingly discussed conceptually, a rigorous benchmarking methodology to quantify them is still missing. Existing frameworks remain limited in addressing the three pillars of modern gripper benchmarking — transferability across robotic embodiments, energy-aware operation for resource-constrained platforms, and reproducible comparison enabling transparent cross-platform evaluation. In addition, current benchmarks do not provide standardized reporting conventions for transfer-related metrics, making it difficult to compare methods across laboratories or robotic platforms.

These limitations motivate the development of a comprehensive and reproducible framework that integrates mechanical, energetic, and functional evaluation of gripper performance across platforms. The goal is not to replace established YCB and NIST procedures but to extend them with complementary metrics that address embodiment transfer and energy efficiency in a unified way. 

The main contributions of this letter are as follows: (1) We introduce the Cross-Embodiment Gripper Benchmark (CEGB), a compact benchmarking suite extending YCB and selected NIST metrics with additional evaluations focusing on cross-platform transfer and energy-aware performance; (2) We present and validate a self-locking reference gripper prototype (RGP) characterized by low energy consumption during holding; (3) We provide open-source data, CAD model, and scripts to support reproducible cross-platform comparison of grippers.

Beyond addressing these gaps, CEGB aims to provide a common language for discussing gripper performance across heterogeneous embodiments. By doing so, it supports more transparent comparison of emerging designs and promotes reproducible evaluation practices across the community.

\section{Cross-Embodiment Gripper Benchmark} \label{sec:utb}

A comprehensive evaluation of a gripper requires benchmarks that are comparable across platforms and reproducible within the community. We build upon the widely adopted YCB Object and Model Set and selected NIST protocols, extending them with complementary metrics that explicitly quantify transferability, energy efficiency, and intent-specific payload. Transferability is understood as the ability of a gripper to maintain its function and performance across distinct robotic embodiments with minimal adaptation, an aspect not addressed in existing standards. The resulting CEGB thus combines established object- and force-based evaluations with additional cross-platform and energy-aware metrics tailored for aerial and manipulators scenarios.

All continuous variables are reported as median, interquartile range (IQR), and 95\% bootstrap confidence intervals, while binomial proportions use Wilson score intervals for consistent statistical reporting. This convention reduces the influence of outliers and non-Gaussian effects, which are common in contact-rich experiments, and facilitates transparent comparison of results across laboratories that adopt the same benchmark. This unified reporting structure ensures that future grippers evaluated under CEGB can be compared directly, regardless of platform, embodiment, or actuation technology.

\subsection{YCB Grasp Success and Metrics}
\label{subsec:ycb}

The YCB Object and Model Set ~\cite{7251504} provides a reproducible baseline for evaluating grasp--hold--release performance and remains the
most widely adopted object set for manipulation benchmarking. We follow established YCB evaluation practices ~\cite{ChenBohgLiu2024}, ~\cite{zhou2023instagraspentirely3dprinted} using a subset of objects $S$ and testing each object $o$ in $k$ canonical poses, with $a$ repeated attempts per pose. The choice of $S$, $k$, and $a$ is driven by a trade-off between experimental time and statistical reliability, and can be adapted by other users of CEGB as long as the configuration is reported together with the results.

For each object $o$ and pose $j$, the grasp success proportion is
\begin{equation}\label{eq:s}
    s_{o,j} = \frac{g_{o,j}}{a}, \qquad j = 1,\dots,k 
\end{equation}
where $g_{o,j}$ is the number of successful attempts. The per-object success rate is
\begin{equation}
    s_o = \frac{1}{k}\sum_{j=1}^{k} s_{o,j}
\end{equation}

Dataset-level summaries are reported as micro- and macro-averages:
\begin{equation}\label{eq:smicro_macro}
    \bar{s}_{\text{micro}} =
    \frac{\sum_{o\in S}\sum_{j} g_{o,j}}
         {\sum_{o\in S}\sum_{j} a},
\qquad
    \bar{s}_{\text{macro}} =
    \frac{1}{|S|}\sum_{o\in S} s_o 
\end{equation}
Here, $\bar{s}_{\text{micro}}$ emphasizes the overall success rate weighted by the number of trials, while $\bar{s}_{\text{macro}}$ assigns equal weight to each object, which is useful when comparing grippers with different object preferences or failure modes.

To quantify uncertainty in these binomial success proportions, we report Wilson score confidence intervals, which offer well-behaved coverage even for small sample sizes and proportions near 0 or 1, and represent the recommended practice in manipulation benchmarking. ~\cite{8950105}. 

A grasp is considered successful if the object is lifted at least 5\,cm within
3\,s and held for 3\,s without slip (10\,s timeout), consistent with prior YCB-based
evaluations. In addition to binary outcomes, we 
record \emph{time-to-lift} and \emph{time-to-release} as complementary continuous 
metrics, reported using median, IQR, and 95\% bootstrap confidence intervals. These temporal quantities provide additional insight into the dynamic behavior of the gripper, for example distinguishing fast but unreliable grasps from slower yet more robust strategies.

\subsection{NIST Benchmark}\label{subsec:nist}

To complement the YCB grasp–success evaluation, we adopt a compact subset
of NIST end-effector metrics~\cite{8950105}, covering timing, force output,
slip resistance, and payload capacity. These quantities provide platform-agnostic, time- and force-based descriptors that capture aspects not addressed by the YCB pass/fail criterion. Within CEGB, the NIST-derived metrics serve as a common reference layer that allows different grippers and embodiments to be compared in terms of fundamental mechanical capability, independently of the specific object set or task.

\paragraph{Grasp Cycle Time}\label{sec:NIST_cycle_time}

The grasp cycle time is defined as
\begin{equation}
    T_{\text{cycle}} = T_{\text{stop}} - T_{\text{start}}
\end{equation}
corresponding to the duration of a complete grasp--hold--release sequence. To ensure comparability, either standardized gauge objects or identical commanded stroke lengths are used across trials ~\cite{241386}. Results follow
the unified statistical convention of median, IQR, and 95\% bootstrap confidence intervals, see Sec. \ref{sec:utb}. 

\paragraph{Grasp Strength}\label{sec:grasp_strength}

Grasp strength is the total normal force exerted by all fingers pressing a rigid
reference object. Let $F_{N_i}$ denote the normal force at finger $i$; then
\begin{equation}
    F_{\text{total}} = \sum_{i=1}^{n} F_{N_i} 
\end{equation}

Peak and plateau values are recorded for each trial and low-pass filtered to
suppress measurement spikes. This metric refines earlier NIST terminology and
ensures unambiguous definition of normal-force summation ~\cite{8950105}. 

\paragraph{Slip Resistance}\label{sec:NIST_slipResistance}
Slip resistance quantifies the tangential load that induces object motion. The effective friction coefficient is
\begin{equation}\label{eq:slip_resistance}
    \mu_{\text{eff}} = \frac{F_{\text{slip}}}{\sum_i N_i}
\end{equation}
where $F_{\text{slip}}$ is the tangential force at slip onset. For designs where an
external torque $T_a$ is applied along a finger of length $L$, we additionally report the dimensionless holding-quality factor
\begin{equation}
    Q_{\text{hold}} = \frac{F_{\text{slip}} L}{T_a}
\end{equation}
supporting different loading modes and test fixtures ~\cite{10.5555/1209344}.
Each condition is repeated multiple times, including worst-case pull directions. In CEGB, $\mu_{\text{eff}}$ and $Q_{\text{hold}}$ together capture how the contact design, surface materials, and finger geometry influence robustness to external disturbances.

\subsection{Transfer Time Benchmark}\label{sec:utb-C}

Transferability refers to the ability of a gripper to operate across different robotic embodiments without mechanical redesign or control recalibration, as emphasized in prior modular and cross-platform end-effector work ~\cite{Salvietti8593877}, ~\cite{Iqbal}, ~\cite{Wang10582538}. While transferability is often discussed qualitatively in the literature, no quantitative metric currently exists to measure the operational effort required
to reuse the same gripper across heterogeneous platforms.

To address this gap, the \emph{Transfer Time Benchmark} measures the duration required to detach the gripper from one platform and reattach it to another. The metric captures practical factors such as connector accessibility,
alignment tolerance, mechanical keying, safety interlocks, and human operator interaction, elements that strongly affect deployability in real-world applications, especially in mobile or aerial manipulation where downtime is critical. Within CEGB, this benchmark explicitly links hardware design choices (e.g., quick-release interfaces, connector placement) to measurable differences in transfer performance.

The average transfer duration is defined as
\begin{equation}\label{eq:average_transfer_duration}
    T_{\text{transfer}} = \frac{1}{N}\sum_{i=1}^{N} T_i 
\end{equation}

where $T_i$ is the duration of the $i$-th attach/detach cycle. Robustness is quantified by the success rate
\begin{equation}\label{eq:robustness_transfer}
    S_{\text{transfer}} = \frac{N_{\text{success}}}{N}
\end{equation}

in which $N_{\text{success}}$ counts transfers completed without mechanical misalignment, electrical-connector faults, or software/communication errors.

\paragraph{Experimental Protocol}

Transfer trials are performed between two representative robotic platforms. To isolate the end-effector exchange process from platform motion, both platforms are placed in a neutral pose, powered off, and mechanically locked. Each trial starts when the operator touches the quick-release interface and ends when the gripper is fully reattached and electrically available to the
host platform. Multiple repetitions are conducted across participant groups (novices, intermediate users, trained operators) to capture human-related
variability, which is known to influence repeatability in modular tool-changing systems. 

\paragraph{Relevance}

The Transfer Time Benchmark quantifies the practical overhead of reusing the same gripper across heterogeneous embodiments. This is particularly important in aerial manipulation, where minimizing downtime directly affects
mission endurance, and in industrial robotics, where rapid transitions between task-specific tools are often required. Unlike existing benchmarks, which evaluate gripper performance on a single platform, this metric captures the operational effort and reliability associated with embodiment transfer. It thereby complements YCB and NIST metrics by providing a direct link between gripper modularity and system-level deployability.

\subsection{Energy Consumption Benchmark}\label{sec:energy_consumption}

Energy consumption is a critical dimension of gripper performance, particularly for aerial and mobile manipulation platforms where onboard
power is limited. Existing benchmarks rarely quantify energy usage in a standardized manner, despite its direct impact on flight endurance, mission duration, and overall system efficiency. We therefore introduce an \emph{Energy Consumption Benchmark} that evaluates the energy required across the main phases of a grasp cycle. The definition is formulated in an actuation-agnostic manner so that the same benchmark can be applied to electrically, pneumatically, or hydraulically actuated grippers without modification of the core metric. By reporting phase-specific energy, CEGB distinguishes between designs that are efficient during motion, during holding, or in both regimes.

The total cycle energy is defined as
\begin{equation}
    E_{\text{cycle}} = E_{\text{grasp}} + E_{\text{hold}} + E_{\text{release}}
\end{equation}
where each component corresponds to a distinct phase of the grasping process. To avoid mechanical irregularities at stroke limits, a power grasp is executed from a fully open configuration to at least 95\% closure.

\paragraph{Energy Components}

$E_{\text{grasp}}$ denotes the energy consumed from the commanded closing motion until the object is securely lifted. $E_{\text{hold}}$ represents the energy used during a prescribed holding interval $t_{\text{hold}}$, and  $E_{\text{release}}$ covers the energy required
from the opening command until the gripper fully opens.

Each energy quantity is computed by integrating the instantaneous input power,

\begin{equation}\label{eq:energy}
    {E = \int P(t)\, \mathrm{d}t}
\end{equation}
where $P(t)$ denotes the instantaneous power supplied to the gripper actuation system. For electrically actuated grippers, this reduces to $P(t) = U(t)\, I(t)$ using measured voltage $U(t)$ and current $I(t)$, whereas pneumatic or hydraulic systems may compute $P(t)$ from their corresponding pressure and flow variables. Thus, the benchmark remains general while allowing platform-specific implementations.


Because holding energy scales with holding duration, we report a standardized 10-second equivalent:
\begin{equation}
    E_{\text{hold10}} = 10\, \overline{P}_{\text{hold}}
\end{equation}
where $\overline{P}_{\text{hold}}$ is the average input power during the stable hold phase. This normalization facilitates comparison across grippers and use cases with different holding durations and duty cycles.

\paragraph{Experimental Protocol}

Measurements are performed under controlled conditions using representative object weights. For each object, the grasp cycles are repeated in staticitally reasonable count, and the three components $E_{\text{grasp}}$, $E_{\text{hold10}}$, and $E_{\text{release}}$ are recorded. Identical approach distances and grasp types are prescribed across repetitions to ensure comparability. Results are reported using the unified statistical convention of median, IQR, and 95\% bootstrap confidence intervals. In addition, sampling the relevant actuation variables at a fixed rate allows the reconstruction of instantaneous power profiles $P(t)$ and their integrals. Such profiles reveal transient behaviors, frictional effects, or compliance-driven responses that are not evident from aggregate energy values alone, and they provide valuable diagnostic insight when comparing different embodiments or actuation systems.

\subsection{Intent-specific Ideal Payload Benchmark}\label{sec:IIPB_theory}

General-purpose benchmarks often fail to capture the conditions under which a gripper is designed to operate most effectively. Many grippers exhibit
distinct performance characteristics depending on their preferred grasp type, object geometry, and structural compliance. To reflect these design-dependent aspects, we introduce the \emph{Intent-Specific Ideal Payload Benchmark} (IIPB), which evaluates the maximum sustainable load under the gripper’s intended operational configuration. In CEGB, IIPB serves as the link between abstract mechanical capability and the concrete use cases for which a gripper has been designed.

Three standardized artifacts are used to represent typical grasping scenarios: (i) a NIST test cylinder, (ii) a rectangular box with PVC-coated contact surfaces, and (iii) a rigid PVC-coated sphere. Each artifact is assigned a
compact symbolic label (C, B, S), enabling concise notation and facilitating comparison across heterogeneous gripper designs, as summarized in Table ~\ref{tab:iipb_codes}.

To account for structural compliance, grippers are classified as rigid (R- rigid) or compliant in one (1S- semicompliant in one axis), two (2S- semicompliant in two axis), or all three (F- fully compliant) principal axes. While compliant fingers enhance adaptability to irregular or fragile objects, they typically reduce the achievable payload; thus, compliance is reported jointly with the ideal payload capacity. This joint reporting encourages designers to explicitly state the trade-offs between adaptability and load capacity, rather than optimizing one at the expense of the other without documentation.

The measurement procedure follows the slip-resistance protocol described in Sec.\ref{subsec:nist}. The artifact is grasped in the gripper’s preferred configuration, and an external pulling force is gradually increased until slippage occurs or
a predefined safety limit is reached. The ideal payload is defined as

\begin{equation}\label{eq:iipb}
    F_{\text{ideal}} = \max(F_{\text{pull}})
\end{equation}

where $F_{\text{pull}}$ is the externally applied tangential force. Each test is repeated statistically reasonable  times for each artifact size, and results are reported using the unified statistical convention (median, IQR, and 95\% bootstrap confidence intervals). When combined with the compliance and grip-type codes in Table~\ref{tab:iipb_codes}, these values enable concise cataloguing of grippers in terms of both their preferred operating regime and their achievable intent-specific payload.

\begin{table}[!ht]
\centering
\caption{Gripper classification codes used for IIPB comparison.}
\label{tab:iipb_codes}
\begin{tabular}{lll}
\hline
\textbf{Parameter} & \textbf{Option} & \textbf{Code} \\
\hline
Compliance Type & Rigid / 1-axis / 2-axis / 3-axis & R / 1S / 2S / F \\
Preferred Grip Type & Wrap / Pinch & W / P \\
Ideal Object Shape & Cylinder / Box / Sphere & C / B / S \\
\hline
\end{tabular}
\end{table}

\section{Experimental Results} \label{sec: experimental_results}

\subsection{YCB Evaluation}

To quantify the baseline grasping ability, we applied the YCB Grasp Success approach using a constant approach trajectory and fixed grasping speed. For each object--pose combination, $a = \text{5}$ attempts were performed, and per-object success rates were computed according to Eq.~\ref{eq:s}--~\ref{eq:smicro_macro}. The RGP achieved a micro-average success rate of $\text{39.5}\% (95\% CI [37.35 , 41.55])$ and a macro-average success rate of $\text{38.7}\% (95\% CI [22.43 , 57.47])$. Spherical and cylindrical objects exceeded $\text{63}\%$ micro-average success rate, while thin and flat objects remained below $\text{1.7}\%$ micro-average success rate and articulated object reached above $\text{80}\%$ micro-average success rate.

\begin{figure}[!ht]
    \centerline{\includegraphics[width=6.5cm]{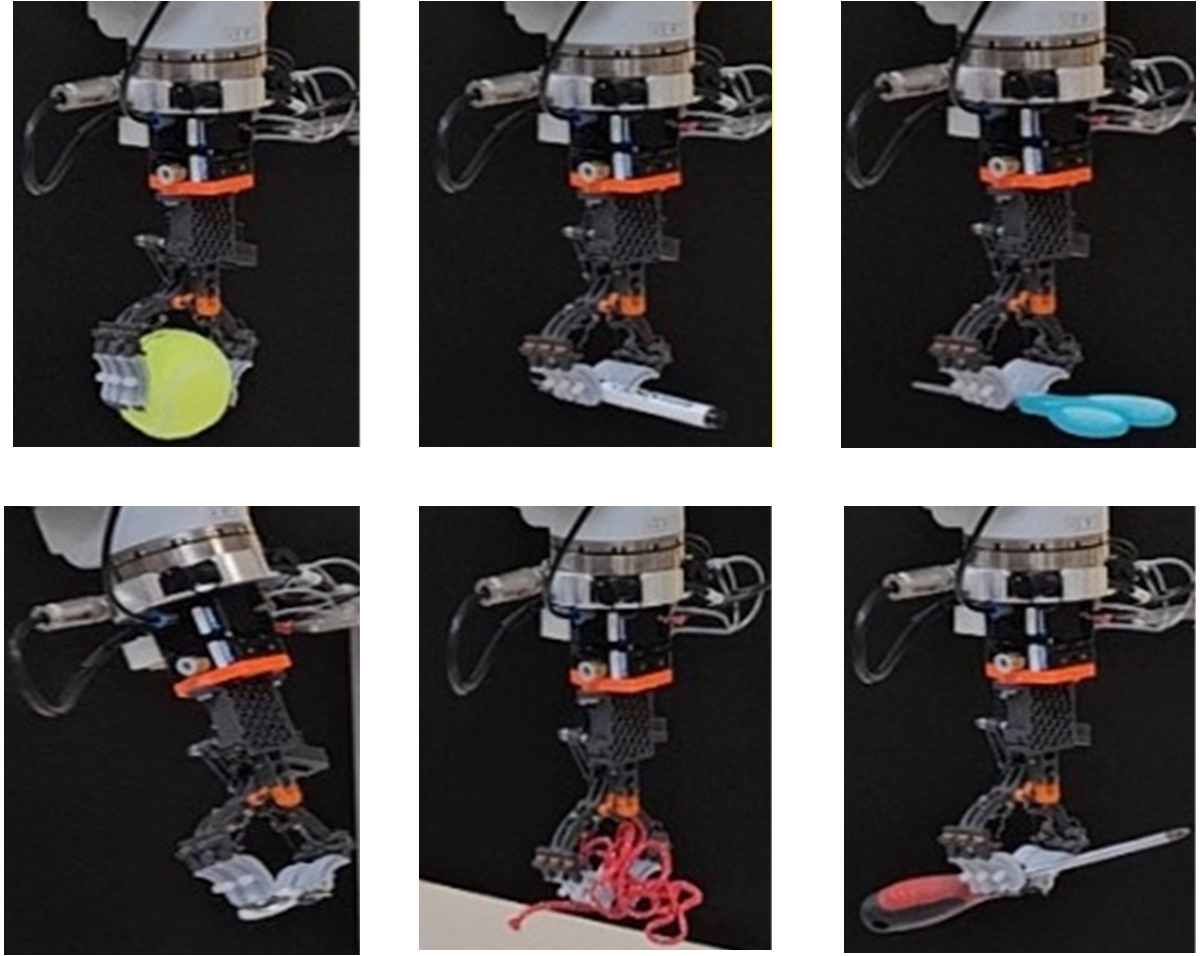}}
    \caption{Reference gripper prototype performance metric results based on the YCB benchmark.}
    \label{fig:gripper_prototype_performance_metric}
\end{figure}

These outcomes are consistent with the gripper’s mechanical structure: high-friction pads and passive finger compliance support adaptation to
curved and irregular geometries, whereas planar artifacts remain challenging due to the fixed fingernail geometry and the absence of a universal picking motion. Fig. \ref{fig:gripper_prototype_performance_metric} illustrates representative examples, including successful grasps of articulated objects by our RGP.

Large or unevenly distributed masses also reduced performance. The main contributors were (i) local pad deformation leading to slippage and (ii)
torsional stresses in the compliant joints causing suboptimal finger alignment. These observations suggest that future iterations may benefit from semi-flexible fingernails to improve ground-level picking.

Across repeated trials, the gripper exhibited reproducible temporal behavior. 
In summary, the YCB evaluation confirms that the RGP achieves robust and reproducible grasping performance across a wide range of object classes, while highlighting expected limitations with thin and heavy items. The
reported success rates with confidence intervals demonstrate performance comparable to established adaptive designs and significantly stronger than
minimal two-finger mechanisms.

\subsection{NIST Evaluation}

The NIST-based evaluation complements the YCB results by providing quantitative measurements of grasp timing, force output, and slip resistance, following the standardized metrics defined in Sec.~\ref{subsec:nist}. Three metrics were evaluated: grasp cycle time, grasp strength, and slip resistance. Each experiment was repeated at least $N=\text{10}$ times under identical conditions, and results are reported as median values with 95\% bootstrap confidence intervals.

\paragraph{Grasp Cycle Time}

The grasp cycle time was evaluated using standard cylindrical artifacts of 50\,mm and 80\,mm diameter, following the NIST procedure described in Sec.~\ref{subsec:nist}. Each cycle began at the commanded approach ($T_{\text{start}}$) and ended once the gripper returned to its fully open
state ($T_{\text{stop}}$). The resulting cycle time is therefore $T_{\text{cycle}} = T_{\text{stop}} - T_{\text{start}}$.

\begin{figure}[!ht]
    \centering
    \includegraphics[width=6cm]{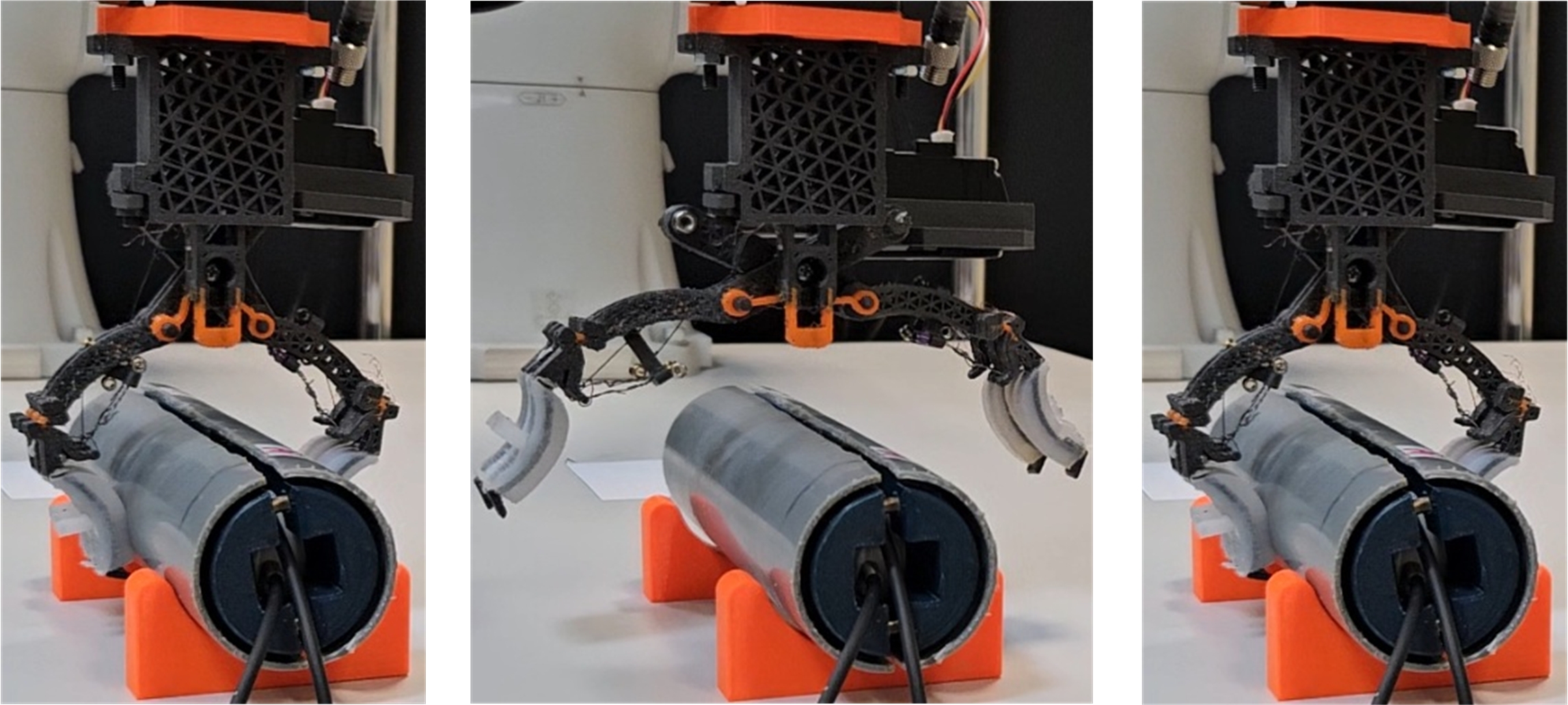}
    \caption{Determination of the reference gripper prototype grasp cycle time following the NIST procedure.}
    \label{fig:cycle_time}
\end{figure}

Each measurement was repeated $32$ times under identical conditions to ensure reproducibility. 
The grasping sequence is shown in Fig.~\ref{fig:cycle_time}. 
The use of standardized reference artifacts facilitates comparison with other grippers tested under the NIST framework.

The mean cycle times and 95\% confidence intervals are summarized in Table~\ref{tab:cycle_time}. The RGP achieved cycle durations of
3.91\,s for the 50\,mm artifact and 3.23\,s for the 80\,mm artifact, with low variability across repetitions. The slight reduction in cycle time for larger artifacts is consistent with the reduced finger travel required for a power grasp at wider diameters.

\begin{table}[!ht]
\caption{Evaluation of the reference gripper prototype grasp cycle time.}
\centering
\begin{tabular}{@{}cccc@{}}
\hline
\textbf{Artifact [mm]} & \textbf{Type} & \textbf{Orientation [$^{\circ}$]} & \textbf{Cycle time [s]} \\ 
\hline
50 & pinch & 0 & 3.91 [3.88, 3.94] \\
80 & pinch & 0 & 3.23 [3.20, 3.60] \\
\hline
\end{tabular}
\label{tab:cycle_time}
\end{table}

\paragraph{Grasp Strength}

Grasp strength was evaluated with 50\,mm and 80\,mm cylindrical artifacts using the NIST force-measurement protocol, defined in Sec.~\ref{subsec:nist}. For each trial, the normal forces exerted by all contacting fingers ($N_i$) were summed to obtain the total grasp force $F_{\text{total}} = \sum_i N_i$.

As shown in  Table~\ref{tab:grasp_strength}, the RGP generated 9.79\,N on the 50\,mm artifact and 8.18\,N on the 80\,mm artifact.
The decrease in force for larger diameters is expected, as finger curvature reduces local contact pressure and the gripping force is not constant through its stroke.

\begin{table}[!ht]
    \centering
    \caption{Evaluation of the reference gripper prototype grasping strength.}
    \label{tab:grasp_strength}
    \begin{tabular}{lccc}
    \hline
    \textbf{Artifact} & \textbf{Grasp Type} & \textbf{Mean [N]} & \textbf{95\% CI [N]} \\
    \hline
    50~mm & pinch & 9.79 & [9.76, 9.82] \\
    80~mm & pinch & 8.18 & [8.15, 8.20] \\
    \hline
    \end{tabular}
\end{table}

\paragraph{Slip resistance}\label{sec:slip_resistance_exp}

The slip resistance test quantifies the gripper’s ability to withstand externally applied tangential loads without object loss, following the NIST methodology defined in Sec.~\ref{subsec:nist}. 
Four cylindrical artifacts with diameters of 32\,mm, 50\,mm, 75\,mm, and 100\,mm were used. 
Each artifact was grasped in a wrap configuration at an orientation of 90°, while the robotic arm applied a constant translational motion along the $y$-axis at a speed of 150\,mm/min. 
A calibrated force sensor recorded the tangential resistance force $F_{\text{slip}}$ until the onset of motion. 
According to Eq.~\ref{eq:slip_resistance}, the effective friction coefficient was evaluated as 
$\mu_{\text{eff}} = F_{\text{slip}} / \sum_i F_{N_i}$, 
where $F_{N_i}$ denotes the normal forces measured during grasping. 
$10$ repetitions were performed for each artifact to ensure statistical reliability, and the mean values with 95\% confidence intervals are reported in Table~\ref{tab:slip}.

\begin{figure}[!ht]
\centering
\includegraphics[width=7.5cm]{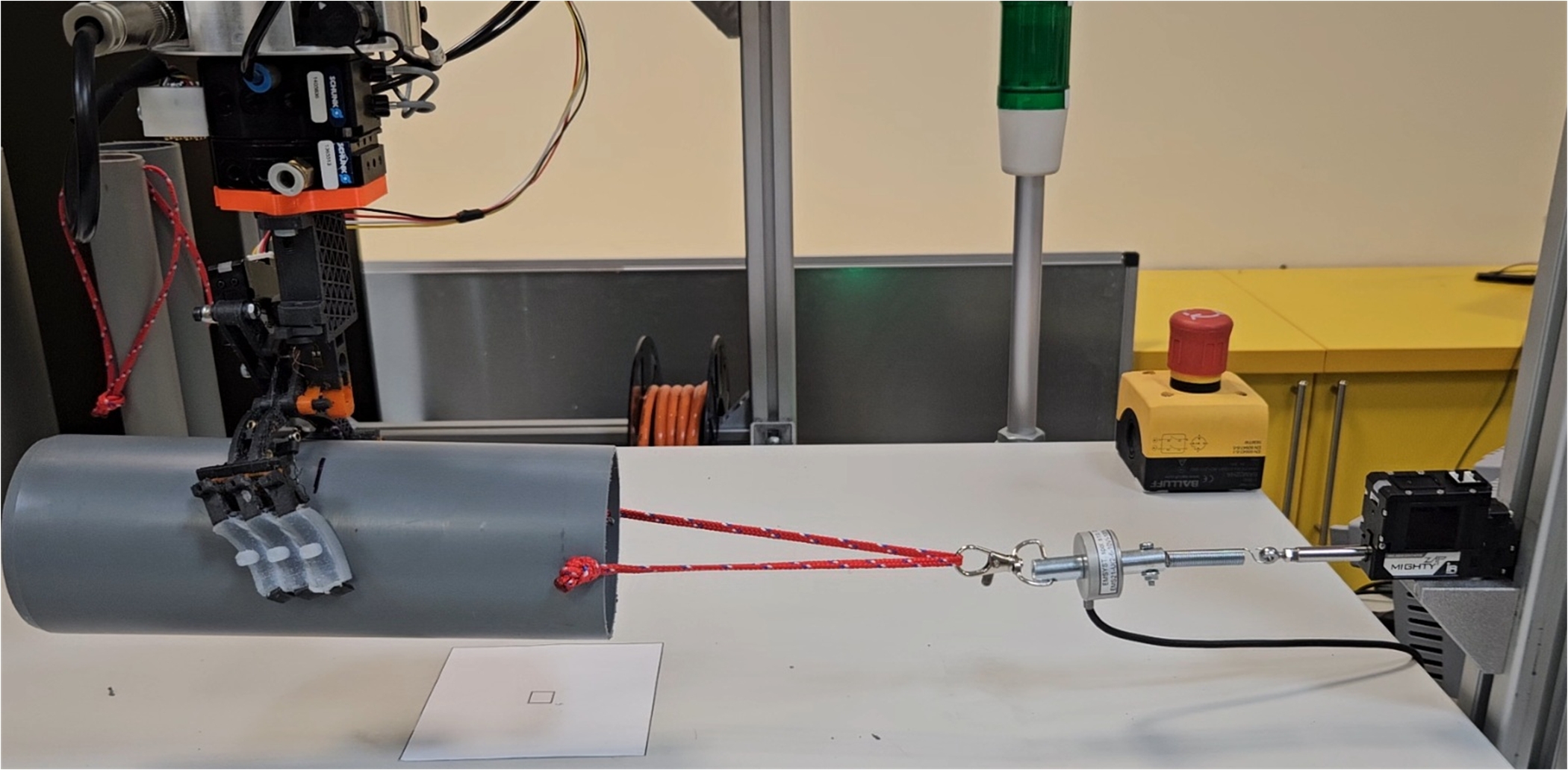}
\caption{Experimental setup for the slip-resistance test based on the NIST methodology.}
\label{fig:slip_test}
\end{figure}

\begin{table}[!ht]
\centering
\caption{Evaluation of reference gripper prototype slip resistance.}
\label{tab:slip}
\begin{tabular}{cccc}
\hline
\textbf{Artifact [mm]} & \textbf{Grasp Type} & \textbf{Mean [N]} & \textbf{95\% CI [N]} \\ \hline
32 & pinch & 6.28 & [6.17, 6.40] \\
50 & pinch & 5.78 & [5.69, 5.87] \\
75 & pinch & 6.24 & [6.06, 6.41] \\
100 & pinch & 3.75 & [3.67, 3.83] \\ \hline
\end{tabular}
\end{table}

The results show consistent behavior across artifacts of comparable sizes, with a noticeable decrease in slip resistance for the largest diameter (100\,mm). 
This trend is attributed to the non constant gripper force throughout the stroke and local deformation of the compliant pads at wider openings. 
Overall, the gripper maintained stable contact forces across the tested range, confirming its ability to resist tangential loads in alignment with the NIST slip-resistance criteria.

\subsection{Transfer Time Evaluation}

The Transfer Time Benchmark introduced in Sec.~\ref{sec:utb-C} was applied to quantify the practical effort required to reuse the same gripper across distinct robotic platforms. For each trial, the transfer time $T_{\text{transfer}}$ was computed as the mean duration of $5$ attach/detach cycles (Eq.~\ref{eq:average_transfer_duration}), while robustness was captured by the success rate $S_{\text{transfer}}$ (Eq.~\ref{eq:robustness_transfer}).

\begin{figure}[!ht]
    \centerline{\includegraphics[width=6.5cm]{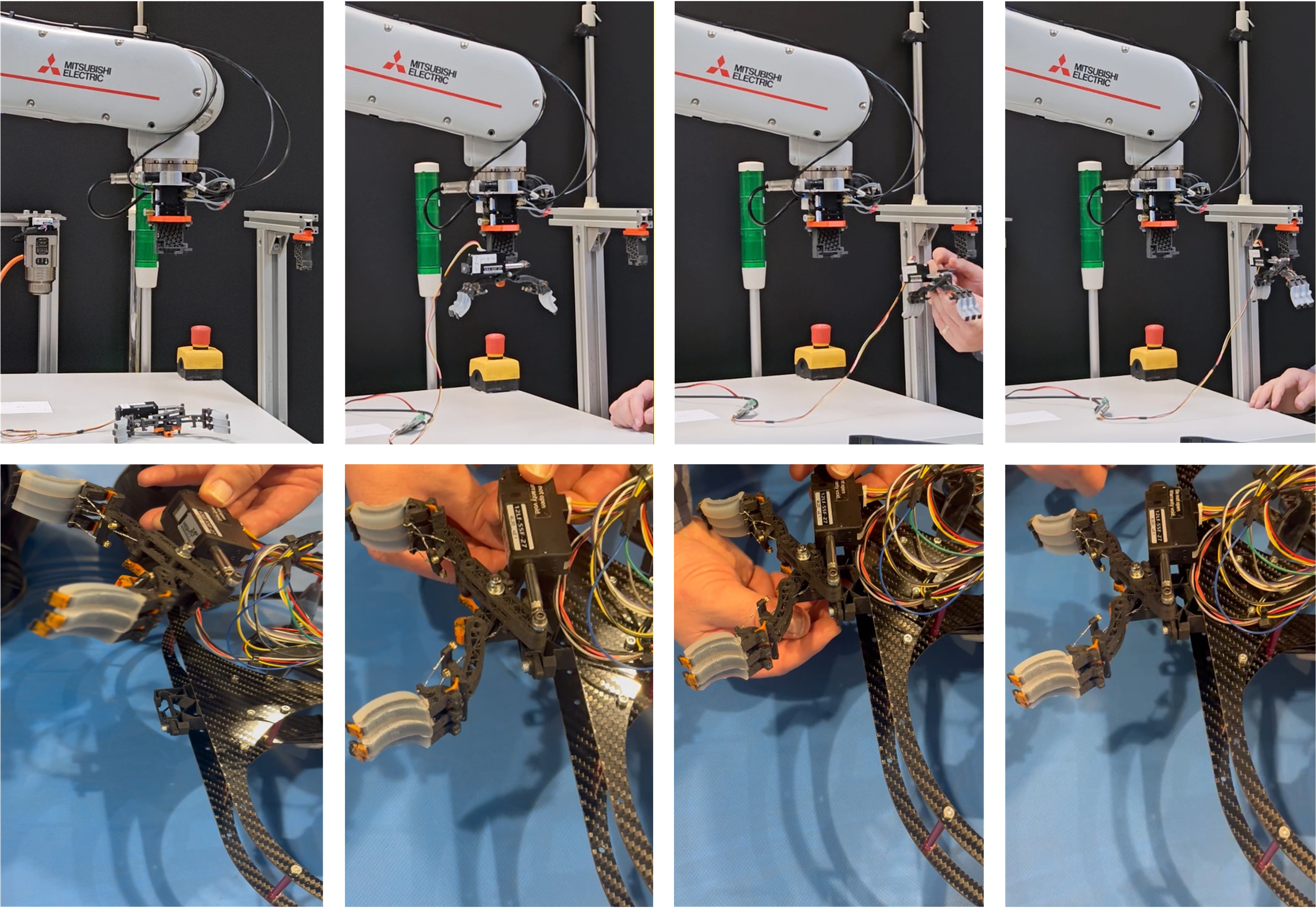}}
    \caption{Experimental sequence for participants to attach and detach the reference gripper prototype.}
    \label{fig:transferability}
\end{figure}

\begin{figure}[!ht]
    \centerline{\includegraphics[width=6.5cm]{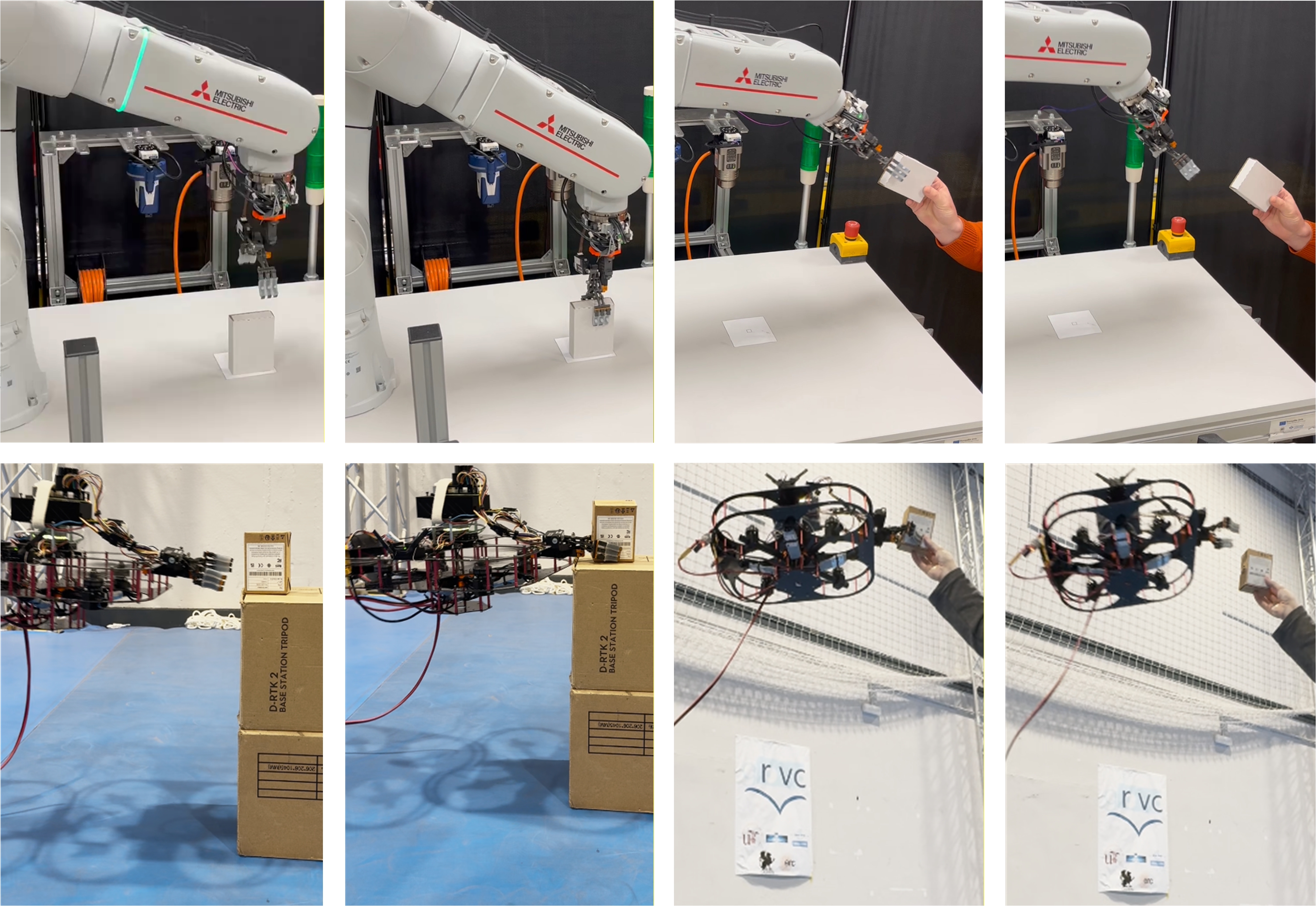}}
    \caption{Transferability to the platforms with parcel delivery task.}
    \label{fig:Transferability to the platform}
\end{figure}

Fig. ~\ref{fig:transferability} illustrates the detach/attach sequence performed by participants. Both platforms were placed in a neutral pose and powered off to isolate the end-effector exchange from any platform. Fig. ~\ref{fig:Transferability to the platform}  shows that after transferring the gripper between the two embodiments, it was immediately deployed in representative pick-and-place tasks on both the industrial manipulator and the UAV, demonstrating functional continuity beyond the attach/detach procedure.

The experiment was repeated across four participant groups (45 in total): bachelor students, master students, colleagues without prior explanation, and experienced researchers involved in this work. 

\begin{figure}[!ht]
    \centerline{\includegraphics[width=7.0cm]{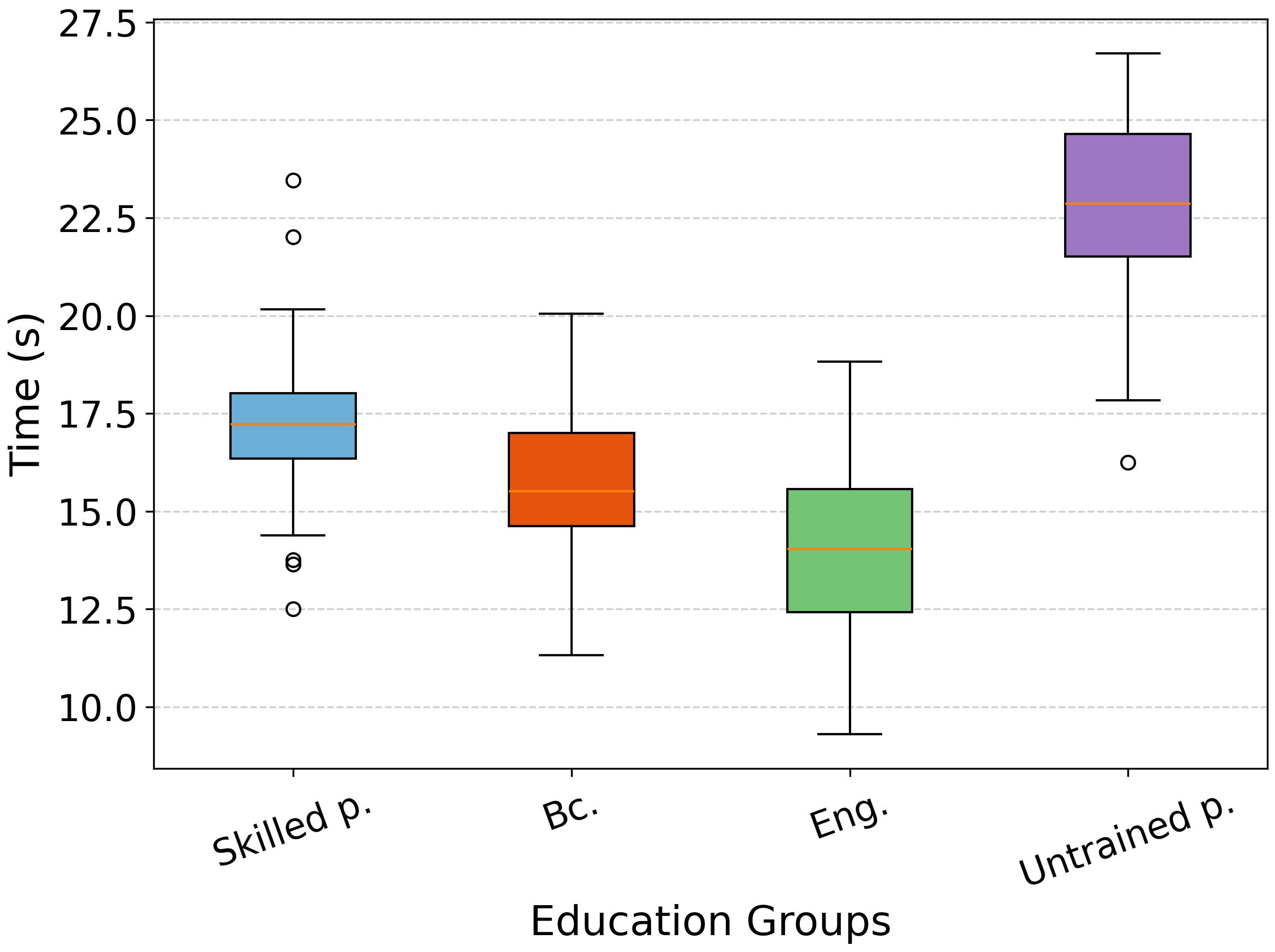}}
    \caption{Transfer times collected from the target participant groups.}
    \label{fig:transfer_time_participants}
\end{figure}

The results are summarized in Fig.~\ref{fig:transfer_time_participants} and reported as median values with IQR 
and 95\% bootstrap confidence intervals. All groups achieved perfect robustness
($S_{\text{transfer}} = 1.0$). The longest transfer durations were observed for
untrained colleagues (22.8\,s), who encountered the mechanism for the first time. Bachelor and master students achieved similar results (16.1\,s and 14.2\,s), reflecting comparable familiarity with robotic hardware. Interestingly, experienced colleagues did not obtain the shortest times (17.6\,s on average); anonymous feedback indicated that some participants handled the mechanism with excessive caution, slowing down the locking step.

Overall, the detachable design enables fast, reliable, and intuitive transfer across robotic embodiments. The fact that inexperienced users performed comparably to trained operators highlights the usability of the mechanism and its relevance in scenarios where minimising downtime is critical, such as aerial or industrial manipulation.

\subsection{Energy Efficiency Evaluation}

The energy efficiency of the gripper was evaluated according to the Energy Consumption Benchmark defined in Sec.~\ref{sec:energy_consumption}. Measurements were performed across the three phases of the grasp cycle—approach and grasping ($E_{\text{grasp}}$), holding ($E_{\text{hold10}}$), and release
($E_{\text{release}}$)—with $10$ repetitions per condition. Instantaneous
voltage $U(t)$ and current $I(t)$ were sampled at 0.1\,kHz, and electrical
energy for each phase was obtained by integrating power over time
(Eq.~\ref{eq:energy}). Each grasping/releasing phase lasted approximately $2\,s$, corresponding to a stable and repeatable motion profile.

\begin{figure}[!ht]
    \centering
    \includegraphics[width=8.5cm]{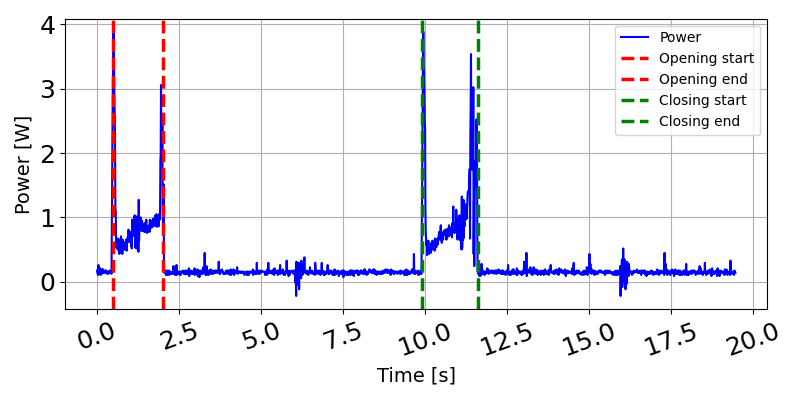}
    \caption{Instantaneous electrical power $P(t) = U(t)I(t)$ measured during the grasp cycle. The plot shows the real-time power profile across the closing, holding, and release phases.}
    \label{fig:energy_phases}
\end{figure}

Fig.~\ref{fig:energy_phases} shows the instantaneous power traces for the three phases of the grasp cycle. Distinct power peaks appear at the onset of closing and at the moment of object contact, 
indicating that the actuator draws the most energy during transient motion, while the steady-state holding phase remains energetically minimal due to the self-locking mechanism. During the holding phase, the self-locking mechanism of RGP prevents back-driving and maintains the grasp with minimal current consumption. This results in a normalized holding energy of $E_{\text{hold10}} 
\approx 1.5$\,J (see Tab.~\ref{tab:energy_test}).

\begin{table}[!ht]
    \centering
    \caption{Evaluation of the reference gripper prototype energy consumption.}
    \label{tab:energy_test}
    \begin{tabular}{lccc}
    \hline
    \textbf{Cycle phase} & \textbf{Mean [J]} & \textbf{95\% CI [J]} \\
    \hline
    Grasping & 2.59 & [2.71, 2.64] \\
    Holding 10s & 1.5 & [1.5, 1.51] \\
    Releasing & 1.91 & [1.89, 1.93] \\
    \hline
    \end{tabular}
\end{table}

The total energy per grasp cycle reached approximately $E_{\text{cycle}}
\approx 6$\,J for a 600\,g object, yielding an energy-to-weight ratio of
0.01J/g for the full cycle and $4.31 \times 10^{-3}$\,J/g for the grasping phase alone. 
Overall, the results confirm that the RGP consumes energy primarily
during active finger motion, with negligible energy required to maintain
a grasp. This validates the assumptions of Sec.~\ref{sec:energy_consumption} and demonstrates the suitability of the self-locking design for energy-limited platforms such as UAVs and other mobile manipulators. Such energy characteristics are particularly advantageous for UAV-based manipulation, where even small reductions in holding power can translate into measurable gains in flight endurance.

\subsection{Intent-specific Ideal Payload Benchmark}

The Intent-Specific Ideal Payload Benchmark (IIPB) introduced in Sec.~\ref{sec:IIPB_theory} was applied to evaluate the gripper under its intended operational conditions. As the proposed design is optimized for lightweight, energy-efficient aerial manipulation, this benchmark quantifies its maximum sustainable load in the preferred grasp configuration.

The gripper exhibits compliance primarily in the $x$- and $y$-axes of the
gripping plane, with limited compliance in the $z$-axis. It predominantly operates in a power–pinch configuration with a reliable grasping range of 40--100\,mm, while wrap grasps are used only in rare situations. The key parameters and compliance classification are summarized in Table~\ref{tab:IIPB_params}.

\begin{table}[!ht]
    \centering
    \caption{Reference gripper prototype parameters and options.}
    \label{tab:IIPB_params}
    \begin{tabular}{lcc}
    \hline
    \textbf{Gripper Parameter} & \textbf{Option} & \textbf{Code} \\
    \hline
    Compliance Type & Compliant in 2 main axes & 2S \\
    Preferred Grip Type & Pinch & P \\
    Ideal Object Shape & Box-type artifacts & B \\
    Gripping Range & 40--100~mm & -- \\
    \hline
    \end{tabular}
\end{table}
\begin{figure}[!ht]
    \centerline{\includegraphics[width=8.0cm]{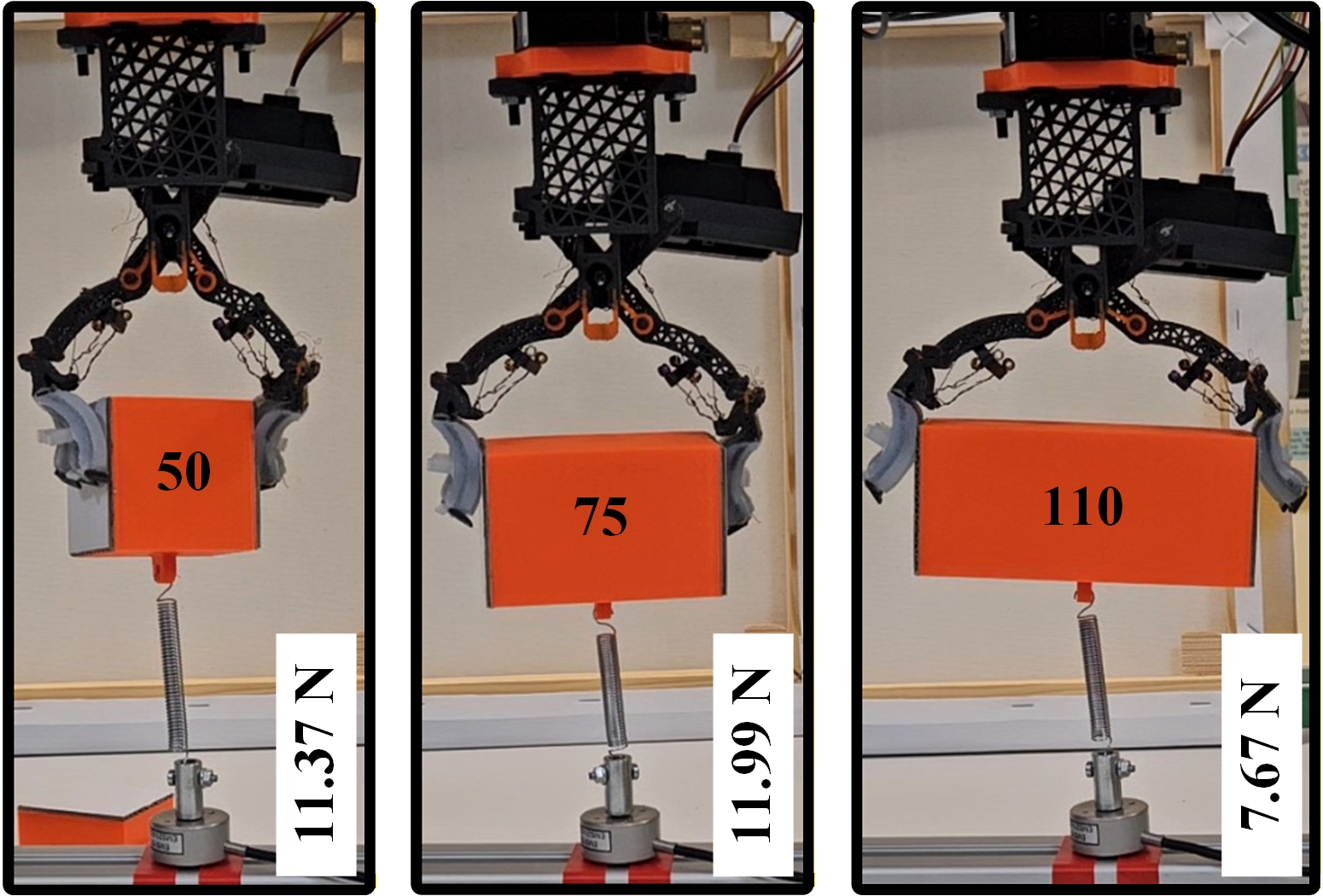}}
    \caption{Experimental setup for the Ideal Payload Benchmark.}
    \label{fig:IIPB_test}
\end{figure}
The measurement procedure follows the slip-resistance protocol in Sec.~\ref{subsec:nist},
adapted for payload evaluation. Three box-type artifacts with gripping surface distances of 50\,mm, 75\,mm, and 100\,mm were tested. One side of the artifact was grasped by the RGP, while the opposite side was attached to a calibrated 100\,N force sensor. During each trial, the robotic arm moved vertically upward at 150\,mm/min until slip or detachment occurred. The maximum recorded pulling force corresponds to the ideal payload as
defined in Eq.~\ref{eq:iipb}, where $F_{\text{pull}}$ denotes the externally applied tangential force. 

Each measurement was repeated $10$ times for each artifact size, and the 
resulting ideal payload distributions are shown in Fig.~\ref{fig:IIPB_test} and summarised in Table~\ref{tab:IIPB_results} using the unified statistical convention (median, interquartile range, and 95\% bootstrap confidence intervals). The highest payload was achieved for the mid-range artifact (75\,mm), while the reduced
load capacity for the 100\,mm artifact is consistent with lower normal pressure,
increased pad deformation at wide openings and non-constant force output throughout the stroke.

\begin{table}[!ht]
    \centering
    \caption{Evaluation of ideal maximum load.}
    \label{tab:IIPB_results}
    \begin{tabular}{lcc}
    \hline
    \textbf{Artifact [mm]} & \textbf{Ideal Max Load [N]} & \textbf{95\% CI [N]} \\
    \hline
    50 & 11.37 & [11.06, 11.69] \\
    75 & 11.99 & [11.59, 12.38] \\
    100 & 7.67 & [7.22, 8.11] \\
    \hline
    \end{tabular}
\end{table}

The IIPB results demonstrate reliable grasping across the RGP’s operational range, confirming its suitability for aerial and mobile tasks that prioritise robustness, energy efficiency, and moderate payload capacity. The IIPB benchmark therefore provides a concise descriptor of a gripper’s intended capability, complementing platform-agnostic metrics such as YCB and NIST with task-specific operational insight.
\section{Conclusion and Future Work} \label{sec: conclusion}

This letter presented the Cross-Embodiment Gripper Benchmark, a reproducible framework that extends YCB and NIST protocols with embodiment-transfer, energy-aware, and intent-specific payload metrics. Using a lightweight self-locking reference gripper prototype, we experimentally validated the framework and demonstrated rapid transferability ($\approx 17.6\,\text{s}$, 100\% success), high holding efficiency ($\approx 1.5\,\text{J}/10\,\text{s}$), and consistent grasp performance (object dependent NIST cycle time $\approx 3.2\,\text{–}\,3.9\,\text{s}$). 

CEGB enables standardized cross-platform comparison of grippers and supports transparent evaluation of mechanical, energetic, and usability-related characteristics across heterogeneous robotic embodiments. Future work will focus on automating the benchmark procedures, extending the evaluations to dynamic aerial and mobile scenarios, and supporting broader community adoption to further advance reproducible, energy-aware, and platform-agnostic benchmarking.






\section*{ACKNOWLEDGMENT}
This work is supported by EUROBIN, the European Robotics and AI Network (Grant agreement 101070596), funded by the European Commission.


\bibliographystyle{IEEEtran}
\bibliography{references}

\end{document}